%% file: emotionpaperarxiv.tex
\documentclass[11pt,conference]{IEEEtran}

\makeatletter
\def\ps@headings{%
\def\@oddhead{\mbox{}\scriptsize\rightmark \hfil \thepage}%
\def\@evenhead{\scriptsize\thepage \hfil \leftmark\mbox{}}%
\def\@oddfoot{}%
\def\@evenfoot{}}

\makeatother

\input{header}
\usepackage{amssymb,color,amsmath}
\usepackage[utf8]{inputenc}
\usepackage[english]{babel}
\usepackage{multirow,multicol}
\usepackage{amsthm}
\usepackage{cite,manfnt}
\usepackage[urlcolor=rltblue,colorlinks=true]{hyperref} %
\usepackage{arydshln}

\usepackage{subfigure}
\usepackage{paralist,color,clock}
\usepackage{graphicx}

\usepackage{algorithm}
\usepackage{algpseudocode}
\newcommand{\algrule}[1][.2pt]{\par\vskip.5\baselineskip\hrule height #1\par\vskip.5\baselineskip}

\usepackage{tikz}
\usetikzlibrary{fit,positioning}

\definecolor{rltblue}{rgb}{0,0,0.75}\usepackage[colorlinks=true,urlcolor=rltblue]{hyperref}

\newcommand{\nix}[1]{}

\begin{document}
\title{Arabic Speech Emotion Recognition  Employing Wav2vec2.0 and HuBERT Based on BAVED Dataset}

\author{Omar Mohamed$^1$ ~~~~~~~~~~~~~Salah A. Aly$^2$\\
$^1$Faculty of Computers and Artificial Intelligence,  Helwan  University, Egypt\\
$^2$Computer Science Section, Faculty of Science,  Fayoum University, Egypt\\
}

\date{}
\maketitle              

\begin{abstract}
Recently, there have been tremendous research outcomes in the fields of speech recognition and natural language processing. This is due to the well-developed multilayers deep learning paradigms such as wav2vec2.0, Wav2vecU, WavBERT, and HuBERT that provide better representation learning and high information capturing.  Such paradigms run on hundreds of unlabeled data, then fine-tuned on a small dataset for specific tasks.
This paper introduces a deep learning constructed emotional recognition model for Arabic speech dialogues. The developed model employs the state of the art audio representations include wav2vec2.0 and HuBERT. The experiment and performance results of our model  overcome the previous known outcomes.

\end{abstract}

\bigskip
\section{Introduction}
\bigskip
Researchers and scientists have used deep learning as Feature extraction in recent decades, utilizing the power of deep learning to capture important features that have resulted in significant improvements in literature and real-world applications.

There are several hitches when pursuing research work in emotion recognition:
\begin{itemize}
  \item emotion detection of humans does not reply on the text or on their speeches, but also in the way they talk to other persons,
  \item the leak of the available dataset sets particularly in Arabic dialogues,
  \item emotion detection does not depend only on one word but also in several words in the contexts,
  \item Some words can be used in different styles, in which they express the speakers’ attitude and emotion.
\end{itemize}

The prosodic properties of human speeches are represented by acoustic features such as  pitch, intensity, duration, and voice quality.

Despite the enormous success contributions in emotion recognition in English datasets, there is still a gab in Arabic dataset and emotion recognition systems utilizes these Arabic datasets. Various Arabic speeches emotion datasets have been proposed in the literature, whether audio or visual, see~\cite{Noh2021,Almahdawi2019,LDC2017S12,Shahin2021b}.

There are well-known several datasets for English, Basic Arabic Vocal Emotions Dataset (BAVED) is a dataset that contains Arabic words spelled in different levels of emotions recorded in an audio/wav format~\cite{Aouf2019dataset}.

The problem of emotion recognition analysis on written text or audio speeches has an immense impact and can affect many sectors in the society and relations between persons. A Multi-Task Learning Emotion recognition system has been proposed to detect hate speeches and offensive languages, see~\cite{Miriam2021} and the references therein. This work can also be extended to detect wav audio hate and offensive speeches.

The paper structure is described as follows.  In Sections~\ref{sec:relatedwork} and~\ref{sec:BAVEDdataset}, we introduce the related work and  note about the BAVED dataset, rspectively. In Section~\ref{sec:model} we describe  the proposed mispronunciation error detection model.  In Section~\ref{sec:simulation} we present simulation studies for the proposed model, and finally, the paper is concluded in Section~\ref{sec:conclusion}.

\bigskip

\section{Related Work}\label{sec:relatedwork}
\bigskip

\begin{figure*}[t]
  \centering
  \includegraphics[width=18cm,height=6cm]{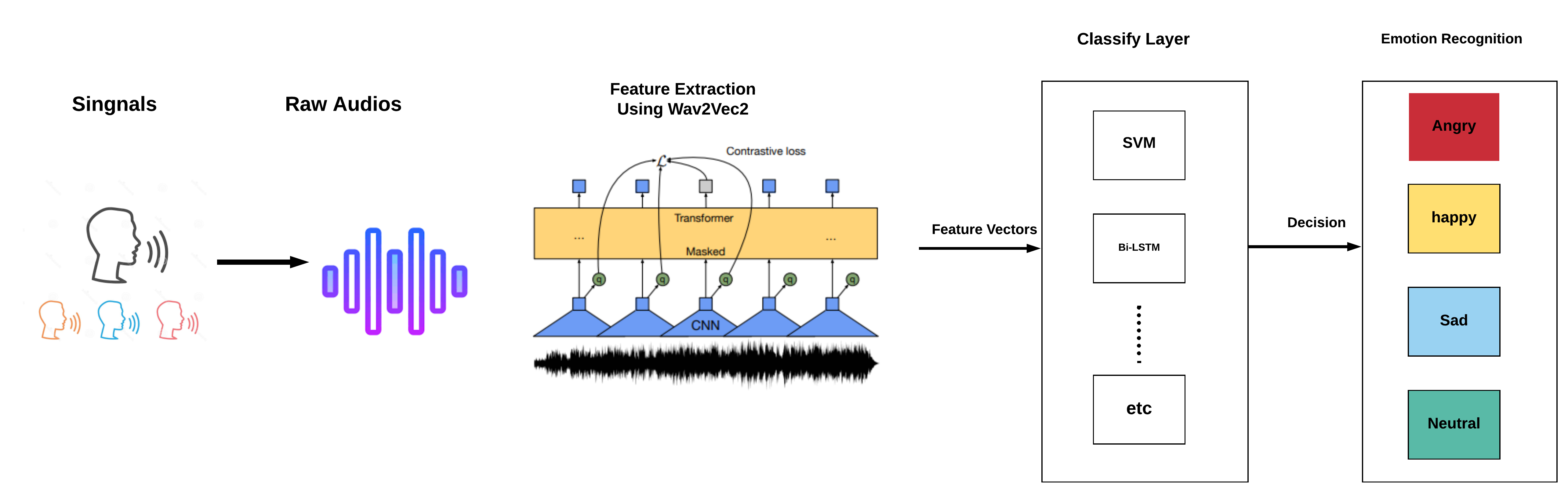}
  \caption{Architecture of the system for emotion recognition of Arabic speeches}\label{fig:model1}
\end{figure*}

Recent tremendous  results in   speech emotion recognition (SER)  have been focused on the utilizations of deep learning and convolutional networks~\cite{Satt2017},~\cite{Lieskovska2021},~\cite{Yenigalla2018},~\cite{Zhang2018},~\cite{Khalil2019},~\cite{Zheng2021}. The task is also investigated in Arabic speech emotion recognition (ASER)  in several recent results~\cite{yasser2019},~\cite{Hifny2019},~\cite{Klaylat2018}.

The problem has a business side effect in the case of customer's satisfactions and given services. For example, the model can measure if the customers are satisfied about certain products in the market. The system can also be used to happiness and sadness of persons by listening many of their conversations.

Attention-based deep neural networks (DNNs)are employed  to give better results than classical neural networks. Klaylat etc. proposed Arabic emotion recognition system based on a data TV news for three labeled emotions:  happy, angry or surprised~\cite{Klaylat2018}. In their work, classification models are proposed that gave approximately $90\%$ Accuracy.

Recent progress in emotion recognition for certain arabic dialect has been also investigated, see for example~\cite{Cherif2021,AbdelHamid2020EgyptianAS}.

KSU Emotions was developed by King Saud University (KSU) and contains approximately five hours of emotional Modern Standard Arabic (MSA) speech from 23 subjects. Speakers were from three countries: Yemen, Saudi Arabia and Syria~\cite{LDC2017S12}.

Klaylat in~\cite{Klaylat2019} described an Arabic dataset that consists of eight videos of live calls between an anchor and a human outside the studio were downloaded from online Arabic talk shows. Each video was then divided into turns: callers and receivers. To label each video, 18 listeners were asked to listen to each video and select whether they perceive a happy, angry or surprised emotion. Silence, laughs and noisy chunks were removed. Every chunk was then automatically divided into 1 sec speech units forming our final corpus composed of 1384 records.
\medskip
\section{Arabic  BAVED Dataset}\label{sec:BAVEDdataset}
Despite the enormous success contributions in emotion recognition in English datasets, there is still  gab in Arabic dataset and emotion recognition systems utilizes these Arabic datasets. Some Arabic speeches emotion datasets have been proposed in the literature, see~\cite{Noh2021,Almahdawi2019,Aouf2019dataset,LDC2017S12,Shaqra2019}. Each dataset has a different set of classes or labels, for example, the Arabic audio acted dataset proposed in~\cite{Meddeb2015} has five labels  (Happiness, Sadness, Neutral, Anger, Fear), and the dataset proposed in~\cite{Klaylat2018} has three classes (Happy, Surprised, and Angry), while the dataset proposed in~\cite{Shaqra2019} has labels (Happy, Sad, Neutral, Angry, Surprise, Disgust).

\begin{figure}[h]
  \centering
  \includegraphics[width=6cm,height=4cm]{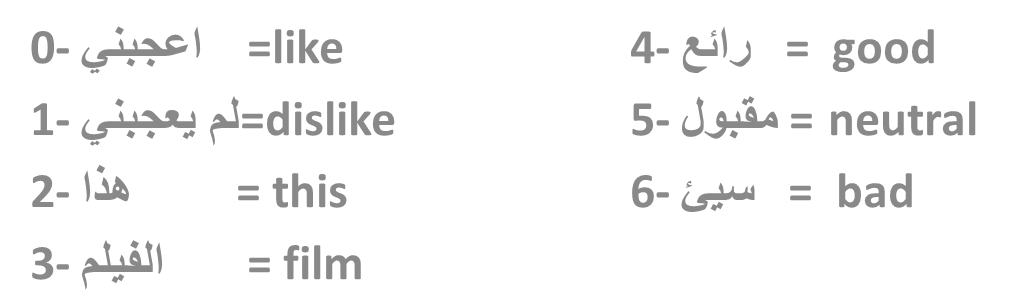}
  \caption{BAVED defined Words~, see\cite{Aouf2019dataset}}\label{fig:bavedwords}
\end{figure}

BAVED dataset is a collection of audio/wav  recorded Arabic words spoken in various expressed emotions~\cite{Aouf2019dataset}.  The  BAVED dataset includes 7 words  given as 0-like, 1-unlike, 2-this, 3-file, 4-good, 5-neutral, and 6-bad. The BAVED dataset word is pronounced in three levels corresponding to the person’s emotions: 0 for low emotion (tired or exhausted), 1 for neutral emotion, and 2 for high emotion positive or negative emotions (happiness, joy, sadness, anger). The dataset contains 1935 recordings that are recorded by 61 speakers (45 males and 16 females). This is a drawback in the dataset that we will investigate in a future work.

\medskip

\section{Methods}\label{sec:model}

Let $\mathcal{A}=\{a_1,a_2,..,a_n\}$  be a set of wav audio signals produced by  several native speakers $L1$. Let $\mathcal{S}=\{s_1,s_2,...,s_n\}$ be a set of words or sentences corresponding to the recordings $\mathcal{A}$, where each sentence (word) $s_j$ corresponding to only one phoneme $a_j$.  Our goal is to detect the emotion of the wav audio signals  $\mathcal{A}$, and check if it is one of the 3 cases in the given dataset BAVED~\cite{Aouf2019dataset}.

\medskip
\subsection{Feature Extraction}

It's relatively simple for a human to understand what's in an image—finding an object, such as a car or a face; classifying a structure as damaged or undamaged; or visually identifying different land cover kinds are all basic tasks. The task is far more complex for machines. To tackle real-world problems, however, it's vital to be able to leverage and automate machine-based feature extraction.
Deep learning is a machine learning technique for detecting features in images. It makes use of a multi-layer neural network, which is a computer system that mimics the functions of the human brain.

Researchers and scientists have used deep learning as Feature extraction in recent decades, utilizing the power of deep learning to capture important features that have resulted in significant improvements in literature and real-world applications.

In the machine learning community, representation learning has evolved into its own area, commonly referred to as Deep Learning or Feature Learning. Although depth is an important aspect of the story, there are many other priors that are intriguing and may be easily captured when the challenge is framed as learning a representation. a more accurate representation allows the model to comprehend the data.
When it comes to representation learning, two recent advanced algorithms produce the state of the art in the field of speech recognition: Wav2vec2.0~c\cite{wav2vec2} and HuBERT~\cite{hsu2021hubert}.

\bigskip

\textbf{Wav2vec2.0:} Wav2vec2.0 is a self-supervised speech representation model that pursues to capture the crucial properties of raw audios by using the power of transformers and Contrastive learning. The wav2vec2.0 training procedure is divided into two phases: i) the model is trained on hundreds of unlabeled data in the first phase, ii) fine-tuned on a small dataset for specific tasks.

The Wav2vec2.0 model consists of:
\begin{itemize}
  \item  convolutional layers that process the raw waveform input to get latent representation – Z,
  \item  transformer layers, creating contextualized  representation – C
	linear projection to output – Y.
We used the pre-trained model Elgeish~\cite{Elgeish2020}, which is Fine-tuned facebook/wav2vec2-large-xlsr-53 on Arabic using the train splits of Common Voice and Arabic Speech Corpus.
\end{itemize}

\bigskip

\textbf{HuBERT:}   Innovative method for self-supervised speech representation learning HuBERT for speech representation learning matches or outperforms SOTA techniques for speech recognition, generation, and compression.
HuBERT learns the structure of spoken input by predicting the proper cluster for masked audio segments using an offline k-means clustering step. By alternating between clustering and prediction processes, HuBERT improves its learnt discrete representations over time. Furthermore, the high quality of HuBERT's learned presentations allows for simple deployment to a wide range of downstream speech applications.

HuBERT uses continuous inputs to train both acoustic and linguistic models. The model must first encode unmasked audio inputs into meaningful continuous latent representations, which correspond to the traditional acoustic modelling problem. Second, the model must capture the long-term temporal relationships between learned representations in order to reduce prediction error. One key finding driving this research is the importance of consistency, not simply correctness, of the k-means mapping from auditory inputs to discrete targets, which allows the model to focus on modelling the sequential structure of input data.

If an early clustering iteration can't tell the difference between the /k/ and /g/ sounds, the prediction loss will learn representations that explain how additional consonant and vowel sounds operate together with this super-cluster to generate words, resulting in a single super-cluster comprising both of them. As a result, the next clustering iteration uses the newly learned representation to produce superior clusters. Our results show that by alternating clustering and prediction phases, representations improve with time.

\bigskip

\subsection{MLP and Bi-LSTM  Classifiers}

After extracting features with wav2vec2.0 and HuBERT, we feed the output into a classifier head:
We utilized MLP Classifier stands for Multi-layer Perception Classifier, which is linked to a Neural Network by its name, and a Bi-LSTM Layer with 50 hidden units, Both classifiers produced results that were close to each other.

\section{Results and Performance Evaluations}\label{sec:simulation}

To test our models, we use three different measurements to evaluate the performance: $F_1$ score, validation loss, and confusion matrix.

\subsection{$F_1$ Score}

We use F-1 score to measure the accuracy of the proposed model.  The reason we use F-1 score is that it gives better measurement for unbalanced data.

\begin{equation}\label{}
  F_1= 2 \frac{precison. recall}{precision + recall}
\end{equation}

\begin{figure}[h]
  \centering
  \includegraphics[width=8.5cm,height=6cm]{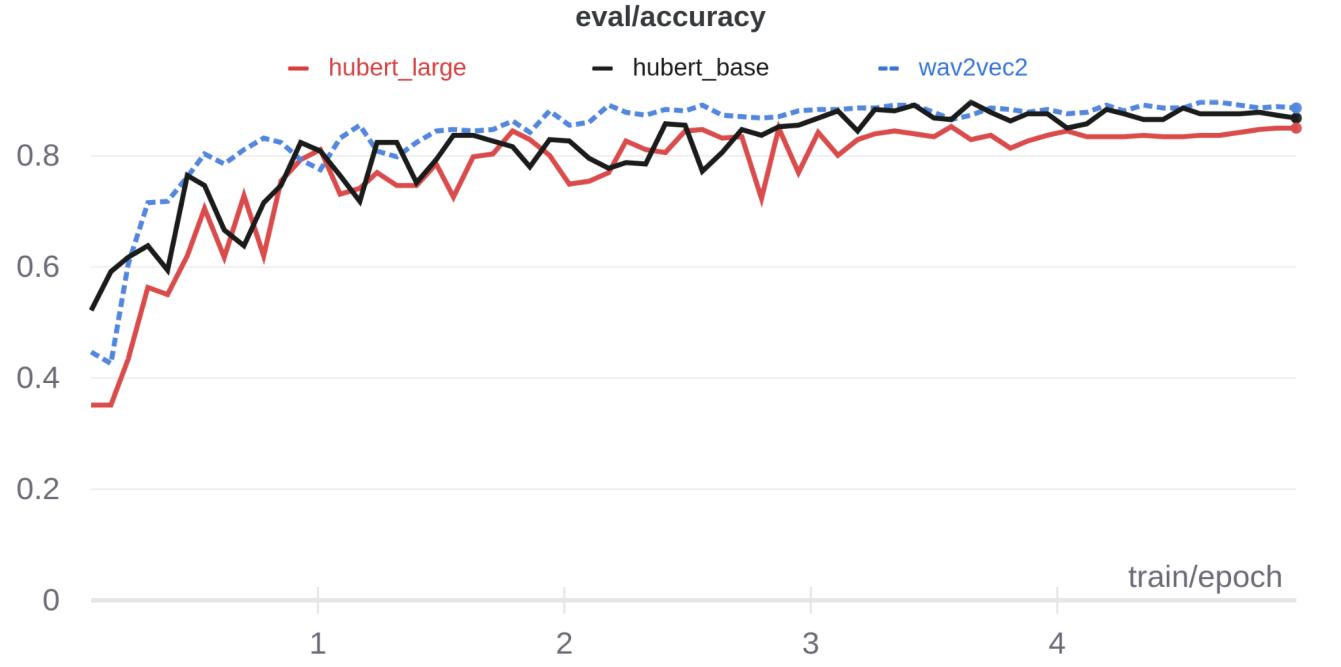}
  \caption{Comprise of the results  and accuracy of the proposed three models}
  \label{fig:f1score}
\end{figure}

Figure~\ref{fig:f1score},  illustrates that  Wav2vec2.0 achieves best accuracy and converge faster than the two other model. HuBERT is unstable during the process of training.

\bigskip

\subsection{Validation Loss and Training Loss}

\begin{figure}[h]
  \centering
  \includegraphics[width=8.5cm,height=7cm]{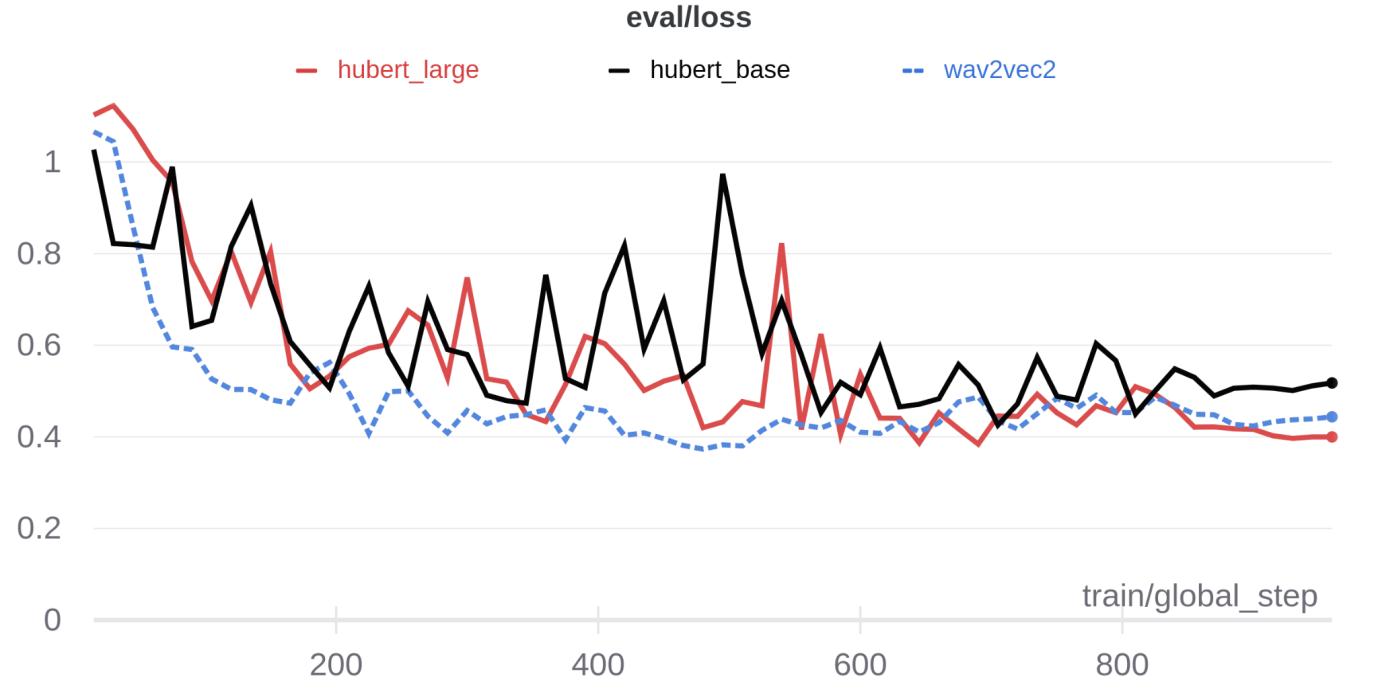}
  \caption{The evaluation loss of the three models.}\label{fig:train_val_loss_hubert_large}
\end{figure}

The models were run for a total of 5 epochs. The Wav2vec2.0 and HuBERT models' outputs are represented in the performance measures used to calculate the categorization problem's performance.  Wav2vec2.0 converges faster than HuBERT model and is more stable during the training phase. As shown in Figure~\ref{fig:train_val_loss_hubert_large}, the train loss of wave2vec2 has the lowest training loss in comparison to the other three models.

Evaluation loss of wav2vec2 has the lowest among the three models , and more stable in the converging process
\begin{figure}[h]
  \centering
  \includegraphics[width=8.5cmm,height=7cm]{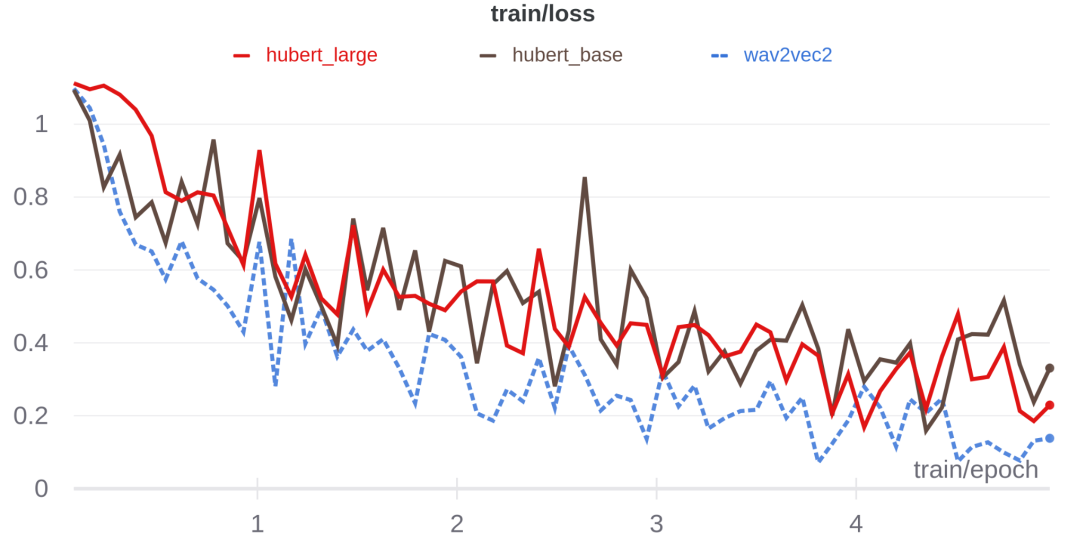}
  \caption{The training loss of the three models.}\label{fig:trainloss_3models}
\end{figure}

\subsection{Confusion Matrix}

\begin{figure*}[t]
  \centering
  \includegraphics[width=5.9cm,height=7cm]{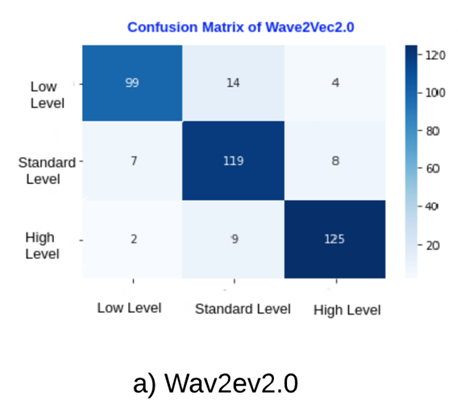}
  \includegraphics[width=5.9cm,height=7cm]{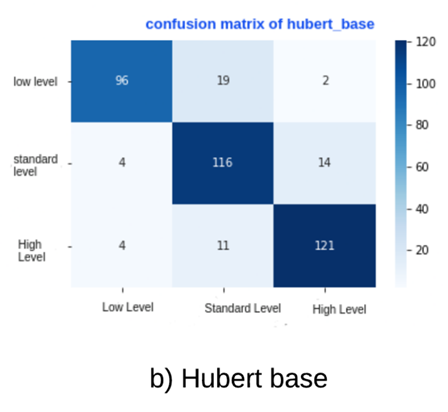}
  \includegraphics[width=5.9cm,height=7cm]{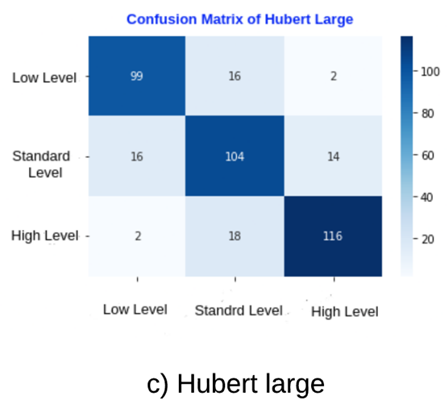}
  \caption{Figures: (a)Results  of Confusion Matrix wav2vec2.0, (b)  Results  of Confusion Matrix HuBEERT base,  (c)    Results  of Confusion Matrix HuBERT large }\label{fig:Confusion-Matrix}
\end{figure*}

Figure~\ref{fig:Confusion-Matrix} displays that the prediction matrix of the start of art models.

\bigskip

Wav2vec2 achieves the best accuracy among the three models.
Despite  HuBERT achieves the best results on the downstream tasks, and captures important feature representation, the wav2vec outperforms for some reasons \begin{inparaenum}
  \item Wav2vec has been trained on Arabic dataset especially Elgeish Pre-trained model,  it has been trained on  Common Voice~\cite{commonvoice2021} and Arabic Speech Corpus~\cite{arabicspeechcorpus2021}.
  \item Both Hubert models (base , large) have been trained on Multi-language Models.
On other tasks HuBERT could outperform the wav2vec2 model.
\end{inparaenum}

\bigskip

As shown in table~\ref{table:result1}, Wav2vec2.0 outperforms HuBERT base and large because Wave2vec2.0's pre-trained model "Elgeish" was trained on Arabic datasets (common voice and Arabic speech corpus), whereas both HuBERT models were trained on multi-language tasks. Despite HuBERT's robustness against noise and ability to capture more information than Wav2vec2.0, it failed this task.

     \begin{table}[h]
            \centering
        \begin{tabular}{|c|c|c|c|}
        \hline
          model &Length & no. records    &accuracy\\
          &&&\\
          \hline
          wav2vec2.0   & 19 Min &1935 & 89  \\
          Hubert Base  & 19 Min &1935 &  87 \\
          Hubert Large & 19 Min &1935 &  84 \\
          \hline
        \end{tabular}
         \caption{Table  of the results  for the proposed three models}
        \label{table:result1}
        \end{table}


The proposed speech emotion procedure is described in Alg.~\ref{alg:cap}. Let $TL$ and $VL$ be the training loss and validation loss, respectively.

\begin{algorithm}
\caption{Training Procedure of AR-Wav2Wav emotion recognition Algorithm}\label{alg:cap}
  \hspace*{\algorithmicindent} \textbf{Input:} {Raw audio sequence A} \\
 \hspace*{\algorithmicindent}  \textbf{Output: }{ Acoustic emotion recognition}
  \algrule
  \begin{algorithmic}[1]
    \Procedure{AR-wave2wav}{$A$}\Comment{Audio recognition}
    \State ${A}\gets []$  and  $V \gets []$
      \vskip 2mm
      \For{i =1 to \text{max-iter}}
    \State Sample a mini-batch of  pairs from ${A}$
    \State \texttt{ $V \gets wav2vec~FeatureExtractor({A_i})$}
    \State Compute $h_o$ as the init. of    A.N.L.
    \State Compute loss for predicted  $y$ and target $Y$
    \State \textbf{Print }   $TL$
    \State \textbf{Print } $VL$
    \EndFor
    \State Print ${a_j}$ emotion recognition
    \EndProcedure
  \end{algorithmic}
\end{algorithm}

\bigskip

\section{Conclusion}\label{sec:conclusion}

We developed the state of the art deep learning models to detect emotions of Arabic speeches.
We implemented these learning models and demonstrated their results on the Arabic BAVED  audio dataset. Several
experiments are performed using wav2vec2 and HuBERT different validation techniques.  The model is successfully
performed by using wav2vec2 and yielded an accuracy of $89\%$. As future work, we plan to extend the proposed method to incorporate more feature sets and increase the size of the dataset for words, sentence, and paragraph recognition.

\bigskip

\section*{Acknowledgement}
This research is partially funded by a grant from the academy of scientific research  and technology (ASRT), 2020, research grant number 6547.

\bigskip

\bibliographystyle{plain} 

\input{emotionref.bbl}

\end{document}

%% file: header.tex
\ifCLASSOPTIONcompsoc

\else
\fi

\ifCLASSOPTIONcompsoc
  \ifCLASSOPTIONconference

  \else

  \fi
\else
\fi